\documentclass{article}

\usepackage{arxiv}

\usepackage[utf8]{inputenc} 
\usepackage[T1]{fontenc}    
\usepackage{hyperref}       
\usepackage{url}            
\usepackage{booktabs}       
\usepackage{multirow}
\usepackage{amsfonts}       
\usepackage{nicefrac}       
\usepackage{microtype}      

\usepackage{graphicx}
\usepackage{subfig}
\usepackage{lipsum}
\usepackage{bbm} 

\usepackage{algorithm}
\usepackage{algorithmic}

\usepackage{booktabs}       
\usepackage{amsfonts}       
\usepackage{nicefrac}       
\usepackage{microtype}      
\usepackage{xcolor}

\usepackage{amsmath}
\usepackage{amsthm}
\usepackage{amssymb}

\usepackage{graphicx}
\def\ALGNAME{LoRa-PGD} 
\def\cifar{CIFAR-10}
\def\imagenet{ImageNet}
\usepackage{comment}
\usepackage{nicematrix}
\usepackage{multirow}
\usepackage{subfig}
\usepackage{lipsum}

\usepackage{mathtools}
\usepackage{enumitem}
\usepackage{tikz}
\usepackage{siunitx}
\usetikzlibrary{patterns}
\usetikzlibrary{arrows.meta}
\usepackage{tcolorbox}
\usepackage{dsfont}

\title{Low-Rank Adversarial PGD Attack}

\author{%
  \textbf{Dayana Savostianova} \\
  School of Mathematics \\
  Gran Sasso Science Institute\\
  L’Aquila (Italy)\\
  \And
  \textbf{Emanuele Zangrando} \\
  School of Mathematics \\
  Gran Sasso Science Institute\\
  L’Aquila (Italy)\\
  \And
  \textbf{Francesco Tudisco} \\
  School of Mathematics \\
  University of Edinburgh (UK)\\
  Gran Sasso Science Institute\\ L'Aquila (Italy)\\
}

\begin{document}

\maketitle
\begin{abstract} 
    Adversarial attacks on deep neural network models have seen rapid development and are extensively used to study the stability of these networks. Among various adversarial strategies, Projected Gradient Descent (PGD) is a widely adopted method in computer vision due to its effectiveness and quick implementation, making it suitable for adversarial training. In this work, we observe that in many cases, the perturbations computed using PGD predominantly affect only a portion of the singular value spectrum of the original image, suggesting that these perturbations are approximately low-rank. Motivated by this observation, we propose a variation of PGD that efficiently computes a low-rank attack. We extensively validate our method on a range of standard models as well as robust models that have undergone adversarial training. Our analysis indicates that the proposed low-rank PGD can be effectively used in adversarial training due to its straightforward and fast implementation coupled with competitive performance. Notably, we find that low-rank PGD often performs comparably to, and sometimes even outperforms, the traditional full-rank PGD attack, while using significantly less memory.
\end{abstract}

\section{Introduction}

Adversarial attacks, characterized by subtle data perturbations that destabilize neural network predictions, have been a topic of significant interest for over a decade \cite{szegedy2013intriguing, goodfellow2014explaining, moosavi2016deepfool, carlini2017towards}. These attacks have evolved into various forms, depending on the knowledge of the model’s architecture (white-box, gray-box, black-box) \cite{vivek2018gray}, the type of data being targeted (graphs, images, text, etc.) \cite{entezari2020all, sun2022adversarial, goodfellow2014explaining, zhang2020adversarial}, and the specific adversarial objectives (targeted, untargeted, defense-oriented) \cite{yuan2019adversarial, madry2017towards}.

While numerous defense strategies aim to broadly stabilize models against adversarial attacks, independent of the specific attack mechanism \cite{cisse2017parseval, galloway2018adversarial, ghiasi2024improving, savostianova2024robust}, the most effective and widely-used defenses focus on adversarial training, where the model is trained to withstand particular attacks \cite{madry2017towards, wang2019improving}. Adversarial training is known for producing robust models efficiently, but its effectiveness hinges on the availability of adversarial attacks that are both potent in degrading model accuracy and efficient in terms of computational resources. However, the most aggressive attacks often require significant computational resources, making them less practical for adversarial training. The projected gradient descent (PGD) attack \cite{madry2017towards} is popular in adversarial training due to its balance between aggressiveness and computational efficiency. 

In this work, we observe that in many cases the perturbations generated by PGD predominantly affect the lower part of the singular value spectrum of input images, indicating that these perturbations are approximately low-rank. Additionally, we find that the size of PGD-generated attacks differs significantly between standard and adversarially trained models when measured by their nuclear norm, which sums the singular values of the attack. This metric provides insight into the frequency profile of the attack when analyzed using the singular value decomposition (SVD) transform, aligning with known frequency profiles observed under discrete Fourier transform (DFT) and discrete cosine transform (DCT) analyses of PGD attacks \cite{yin2019fourier, maiya2021frequency}.

Building on these observations, we introduce \ALGNAME{}, a low-rank variation of PGD designed to compute adversarial attacks with controllable rank. Our results demonstrate that perturbing only a small percentage of the image's singular value spectrum can achieve accuracy degradation comparable to full-rank PGD when the perturbation size is measured using the pixel-wise $l^2$ norm while requiring only a fraction of the memory. Furthermore, when measured using the nuclear norm, these targeted low-rank attacks are significantly more effective than full-rank PGD. This approach not only enhances the effectiveness of adversarial attacks but also reduces the resource requirements associated with generating adversarial examples during training.

\section{Related Work}

Following the seminal contributions of \cite{szegedy2013intriguing,goodfellow2014explaining}, the past decade has witnessed significant efforts to understand the stability properties of neural networks. This has led to the development of various adversarial attacks, categorized broadly into model-based approaches \cite{Poursaeed_generative_Adversarial,laidlaw2019functionaladversarialattacks,xiao2019generatingadversarialexamplesadversarial,Song2018_advpgd} and optimization-based methods \cite{moosavi2016deepfool,goodfellow2014explaining,madry2019deep}. Concurrently, several robustification strategies have been proposed to mitigate these attacks \cite{madry2017towards,bai2021recent}. A key challenge in these approaches is the dependency on the availability of adversarial examples during the training phase, making the efficiency of generating such examples a central concern. Furthermore, the perceptibility of adversarial attacks is another critical factor that has been explored in various studies \cite{croce2019_sparse,qin2019imperceptible}.

Recent research has also begun to focus on the frequency properties of adversarial attacks, making it an emerging area of interest in the community \cite{maiya2021frequency,Luo_2022_CVPR,guo2019lowfrequencyadversarialperturbation,sharma2019effectivenesslowfrequencyperturbations}.

\paragraph{Adversarial Attacks and Training}

The discovery of adversarial examples has catalyzed extensive research on developing robust models. Typically, ensuring robustness against adversarial attacks is computationally intensive, as it often requires the generation of adversarial examples during training \cite{goodfellow2014explaining,xiao2019generatingadversarialexamplesadversarial}. However, adversarial training is not the only method for achieving robustness. Other approaches involve imposing constraints on models to enhance their robustness \cite{leino2021globallyrobustneuralnetworks,savostianova2024robust,zhang2022rethinking,fazlyab2023certified}. The primary distinction between these approaches lies in their trade-offs: adversarial training techniques generally offer higher robustness but are more expensive due to the need for real-time adversarial example generation. In contrast, general robustness techniques may be less efficient in terms of robustness but are consistent across different scenarios. This trade-off highlights the challenge of transferability in adversarial attacks, where adversarially robust models tend to be highly effective against specific attacks but less so in broader contexts.

\paragraph{Low-Rank Structures}

The exploration of low-rank structures has a longstanding tradition in deep learning and numerical analysis. Classical applications range from singular value decomposition (SVD) for image compression to tensor networks used for efficiently representing quantum multi-particle wave functions \cite{Orus_Tensornet2019} and their time integration \cite{Ceruti_ttnintegration,Koch_DLRA}. In machine learning, low-rank methods have been integral to various applications, including recommender systems \cite{Koren_recommender}, latent factor models, and principal component analysis (PCA) \cite{Pearson_PCA,Joliffe_PCA}. More recently, low-rank factorizations have gained attention in deep learning for their ability to provide efficient neural network representations and training methodologies \cite{Idelbayev_2020_CVPR,schotthöfer2022lowranklotteryticketsfinding,Novikov_TensorNN,hu2022lora}. Notably, \cite{hu2022lora} demonstrated the potential for efficiently fine-tuning pretrained large models using small, low-rank additive corrections, further emphasizing the versatility and importance of low-rank structures in modern machine learning.

\section{Low-Rank PGD-style attack}
\subsection{The singular spectrum of PGD attacks}
Let us consider a general setup based on the image classification task. Specifically, let $X \in \mathcal{C} \subseteq \mathbb{R}^{C \times N \times M}$ be an input data tensor with $C$-channels ($C = 3$ for RGB images) of sizes $N \times M$ and $Y \in \mathbb{R}^{D}$ the target tensor. Assume also we have a parametric model $f_\theta: \mathbb{R}^{C \times N \times M} \to \mathbb{R}^{D}$ that has been trained with respect to a loss function $\ell: \mathbb{R}^{D} \times \mathbb{R}^{D} \to \mathbb{R}$. An untargeted adversarial attack on the input data $X$ can be naturally formulated as a perturbation $\delta X^*$ of bounded norm corresponding to the largest induced loss change for the model output, or, in other words, a solution to the constrained optimization problem:
\begin{equation}\label{eq:adv_attack_formulation}
\begin{cases}
    &\delta X^* \in \underset{\delta X \in \mathbb{R}^{C\times N \times M}}{\arg \max} \ell(f_\theta(X+\delta X),Y) \\
    & \|\delta X \|_\omega \leq \tau, \; X+\delta X \in \mathcal{C}
\end{cases},
\end{equation}
where $\tau$ is referred to as the perturbation norm or perturbation budget and is used to control the size of the attack. Small (in terms of $\| X\|_\omega$ ) values of $\tau$ define imperceptible attacks.
Note that, in practical applications and for general neural networks, the optimization problem formulated in \eqref{eq:adv_attack_formulation} is non-convex but the constraint is compact, guaranteeing the existence of at least one solution.
The choice of the norm \( \| \cdot \|_\omega \) significantly influences the nature of the attack \( \delta X^* \), with common choices being \( \| \cdot \|_\infty \) (entry-wise maximum) and \( \| \cdot \|_2 \) (entry-wise Euclidean norm). Importantly, the norm \( \| \cdot \|_\omega \) may directly affect both the effectiveness and the perceptibility of the attack, see e.g.\ \cite{beerens2023adversarialink, sharif2018suitability}. Although all norms are mathematically equivalent, they measure different aspects of the perturbation, leading to variations in the corresponding controlling constants. Consequently, an attack that appears small under one norm may still be perceptible to the human eye if it is large under another norm, and vice versa.

\begin{figure*}[t]
\includegraphics[width=0.49\linewidth]{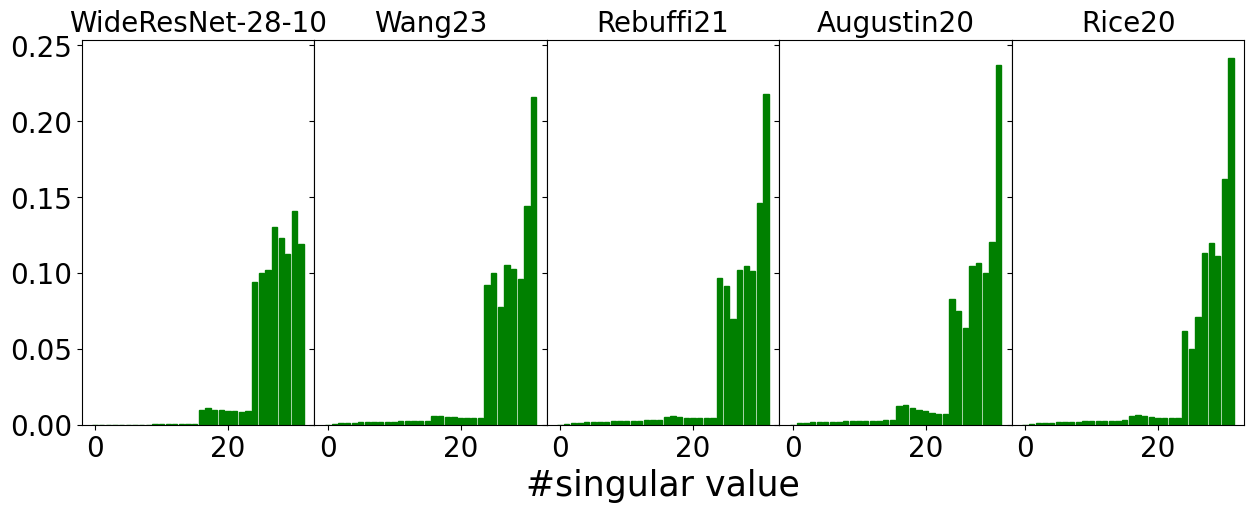}\hfill 
\includegraphics[width=0.49\linewidth]{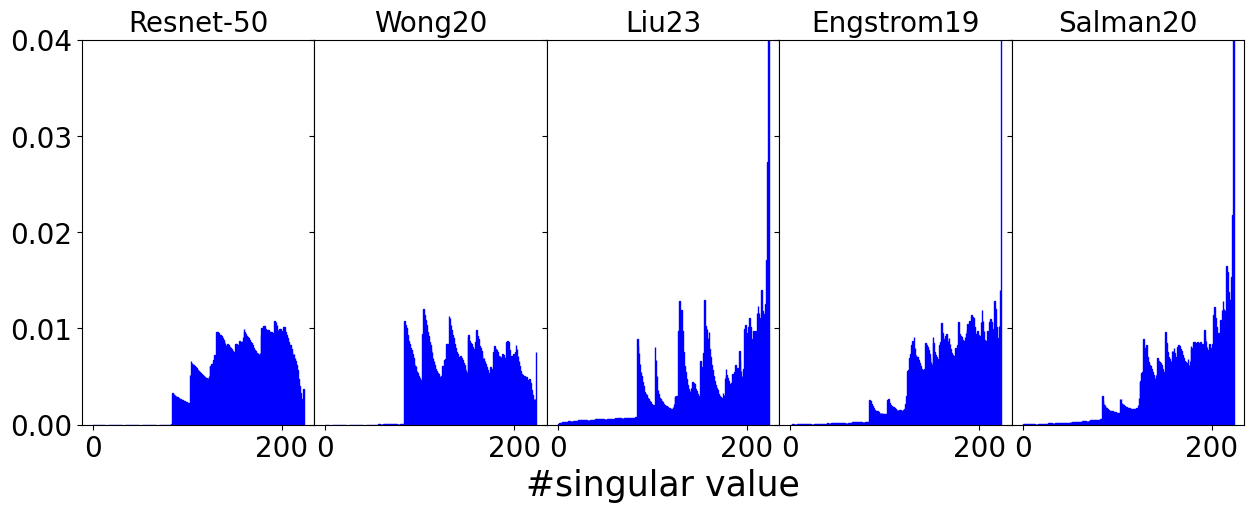}
\caption{Relative magnitude of singular value change for the PGD-attacked images averaged on 5000 images from \cifar{} (in green) and \imagenet{} dataset (in blue), for WideResNet-28-10, Resnet-50, as well as several robust (adversarially trained)  models described in the experiment section}
\label{fig:rel_sing}
\end{figure*}

Although obtaining a global optimizer \(\delta X^*\) for adversarial attacks is challenging due to the problem's inherent complexity, effective approximations can be efficiently derived. One of the most widely adopted and computationally efficient methods is the Projected Gradient Descent (PGD) scheme \cite{madry2017towards}. This method approaches the problem \eqref{eq:adv_attack_formulation} by iteratively taking steps in the direction of the (positive) gradient of the loss, projected onto the unit sphere.

Traditionally, PGD is applied to the flattened image tensor $(X)$, treated as a vector of pixels and channels. However, this flattening process overlooks the spectral structure of the image, thereby obscuring potentially important information in the attack. By reshaping the PGD attack back into a tensor and analyzing the singular values of its channels, we observed an intriguing low-rank pattern. Specifically, we observed that the PGD attack frequently influences only a subset of the singular values of the original image, implying that these perturbations are approximately low-rank. Figure \ref{fig:rel_sing} illustrates this phenomenon, showing the normalized relative change in singular values for 5000 PGD-attacked images from the \cifar{} and \imagenet{} dataset on the models described in \ref{Exp}.

Building on this observation, in the next subsection we propose a variation of the PGD attack that computes low-rank perturbations efficiently, with comparable attack's execution time. When evaluated using the conventional pixel-wise \(l^2\) norm, these low-rank attacks are as effective as the standard full-rank PGD attacks but require significantly lower rank (and thus reduced memory). Furthermore, when assessed using the spectral-focused nuclear norm \(\|\cdot\|_*\), the low-rank attacks demonstrate substantially better performance, still at a fraction of the memory cost.

\subsection{\ALGNAME{}: Low-rank PGD attack}

Similar to what is done in the context of parameter-efficient adaptation of large language models \cite{hu2022lora}, we consider here a variation of PGD that directly searches for a low-rank structured attack. 

Recall that PGD attack is a way to produce a fast approximate solution to \eqref{eq:adv_attack_formulation}. Precisely, the final attack is produced as the limit point of a projected gradient ascent sequence 
\[
\delta X_{k+1} = \Pi\Bigl(\delta X_{k}+ \tau \frac{\nabla_{\delta X}\ell}{\|\nabla_{\delta X}\ell \|}\Bigr), 
\] 
where $\Pi$ is the projection onto the constraint set $\mathcal{C}-X$. In typical applications, $\mathcal C$ just describes the maximal and minimal values each RGB channel can take (thus, $\mathcal C$ is an hypercube) and the projection with respect to the $l^2$ metric can be computed in closed form, and it is given by the $\Pi = \mathrm{clamp}$ function. The final PGD attack is ideally $\delta X^*$, which is a potential maximizer of the loss function on the constraint $\mathcal{C}$.

A low-rank attack can be formulated by looking at a rank-constrained version of \eqref{eq:adv_attack_formulation}:
\begin{equation}
\begin{cases}
    &\delta X^* \in \underset{\delta X \in \mathbb{R}^{C\times N \times M}}{\arg \max} \ell(f_\theta(X+\delta X),Y) \\
    & \|\delta X \|_\omega \leq \tau, X+\delta X \in \mathcal{C}, \mathrm{rank}(\delta) \leq r
\end{cases}
\end{equation}
Given the complex geometrical nature of the rank constraint, one simple way to ensure it is to superimpose a low-rank representation for the perturbation, i.e.
find the solution of the following optimization problem:
\begin{equation}\label{eq:lr_parameterized_adv_attack_formulation}
\begin{cases}
    &(\delta U^*, \delta V^*) \in \underset{\substack{\delta U \in \mathbb{R}^{C\times N \times r},\\ \delta V \in \mathbb{R}^{C\times r \times M}}}{\arg \max} \ell(f_\theta(X+\delta U \otimes_C \delta V),Y) \\
    & \| \delta U \otimes_C \delta V\|_\omega \leq \tau, \; X + \delta U \otimes_C \delta V \in \mathcal{C}
\end{cases}
\end{equation}
where $\otimes_C $ denoted a channel-wise tensor multiplication, 
$$
(A \otimes_C B)_{cij} = \sum_{k=1}^r A_{cik}B_{ckj}\, .
$$
Given the computational prohibitive cost for large models, the classical way to obtain an approximation of a minimizer is through the use of first-order methods. An approximation $(\delta U^*, \delta V^*)$ can be obtained in a PGD-like fashion by computing several steps of  gradient ascend with respect to the smaller variables $\delta U$ and $\delta V$.

In Algorithm \ref{alg:main_algorithm} we present a detailed pseudocode of the resulting \ALGNAME{} attack.  We specify that the $\mathrm{normalize}$ function is just division by the norm, i.e. $\mathrm{normalize}(z) = \frac{z}{\|z\|}$ and the $\mathrm{clamp}$ function is the one defined in {\texttt{torch}}. The former is performed image-wise and the latter -- entry-wise.
Moreover, we notice that the overall scheme proposed in Algorithm \ref{alg:main_algorithm}, when assembled all together in one step,
resembles a simple PGD iteration. In fact, using the update steps for 
$\delta U$ and $\delta V$ in Algorithm \ref{alg:main_algorithm}, we obtain:

\begin{equation}
 \label{eq:ukvk_grad}  
\delta U_{k+1} = \delta U_k+\frac{\nabla_{\delta U} \ell}{\|\nabla_{\delta U} \ell \|}, \quad 
\delta V_{k+1} = \delta V_k+\frac{\nabla_{\delta V} \ell}{\|\nabla_{\delta V} \ell\|}
\end{equation}
where the gradient is evaluated on the new image, perturbed as follows
\[
X = \mathrm{clamp}\Big(X+\tau \, \mathrm{normalize}(\delta U_{k+1}\otimes_C \delta V_{k+1})\Big)
\]

Similar to PGD iteration, $\mathrm{clamp}$ plays the role of the projection (projection on the data constraint $\mathcal{C}-X$). In particular, note that the gradient scheme \eqref{eq:ukvk_grad} remains computationally close to a basic PGD computation since gradients $\nabla_{\delta U} \ell$ and $\nabla_{\delta V} \ell$ are one tensor multiplication away from the computation of the PGD-gradient $\nabla_{\delta X} \ell$.

\begin{algorithm}[tb]
\caption{Pseudocode of (\textbf{\ALGNAME{}}): low-rank gradient attack }
   \label{alg:main_algorithm}
   \begin{algorithmic}
    \STATE Given $X \in \mathbb{R}^{B\times C\times N\times M}$  input tensor, with $Y$ -- target vector, $r$ -- chosen rank, $\tau$ -- perturbation size.
    \STATE $\delta U \leftarrow \mathrm{normalize}(\mathrm{init}([B, C, N, r])) $\;
    \STATE $\delta V \leftarrow \mathrm{normalize}(\mathrm{init}([B, C, r, M])) $\;
        \FOR{step in steps}
            \STATE $\widetilde{X} = \mathrm{clamp}(X + \tau \cdot\mathrm{normalize}(\delta U \otimes_C \delta V), 0, 1)$\;
            \STATE $\delta U += \mathrm{normalize}(\nabla_{\delta U} \ell(f_{\theta}(\widetilde{X}),Y))$\;
            \STATE $\delta V += \mathrm{normalize}(\nabla_{\delta V} \ell(f_{\theta}(\widetilde{X}),Y))$\;
        \ENDFOR
    \STATE $\delta = \tau\cdot\mathrm{normalize}(\delta U \otimes_C \delta V)$\; 
    \STATE $\widetilde{X} = \mathrm{clamp}(X + \delta,0,1)$\; 
   \end{algorithmic}

\end{algorithm}

\paragraph{Nuclear norm and low-rank matrices}
The perturbation budget $\tau$ is typically measured in terms of vector $l^p$ norms, the $l^\infty$ and the $l^2$ norm predominantly, enforcing an upper bound on the size of the perturbation pixel-wise.  However, in order to measure the size of the perturbation from the rank point of view, we additionally consider here the nuclear norm which sums the singular values of the perturbation. Precisely, if we let  $R=\min\left(N,M\right)$, then we set
\begin{equation}\label{eq:nuclear_norm}
    \| X\|_*  = \frac{1}{C}\sum_{i}^{C}\sum_{j}^{R}\sigma_{j}(X_{i,:,:}) = \frac{1}{C}\sum_{i}^{C} \|X_{i,:,:} \|_{*} \, 
\end{equation}
Notice that this coincides with the $l^1$ norm of the singular value vector. Thus, it represents a convex relaxation of the $l^0$ Shatten norm and its usage is analog to the use of the $l^1$ norm as a convex surrogate for the $l^0$ norm in compressed-sensing problems or robust PCA \cite{Rauhut_compressed_sensing,candes2009robustprincipalcomponentanalysis}.

While a small perturbation in $l^p$ norm is bounding the size of the perturbation in the pixels domain, a small perturbation in the nuclear norm is bounding the size of the perturbation in the spectral domain. This can be interpreted as a perturbation targeting the frequencies (the singular values) rather than pixel values, with the perturbation $U \otimes_C V$ attacking specific most relevant singular values of the corresponding image.

\section{Experiments}
\label{Exp}

To provide a fair comparison, RobustBench \cite{croce2021robustbench} was used for pre-trained models, and {\texttt{adversarial-attacks-pytorch}} \cite{kim2020torchattacks} library was used as a basis for the attack comparison.  
We recall that in this section, we will always refer to the $l^2$ norm $\|\cdot \|_{2}$ meaning the entrywise $l^2$ norm, i.e. the Frobenius norm for matrices.
Implementation of the experiments is included in the supplementary material.

\paragraph{Comparison with other methods}

In the experimental section, we compare \ALGNAME{} with the standard implementation of the PGD attack \cite{madry2017towards}. Note that an alternative obvious baseline would be computing a low-rank attack by further projecting the full PGD attack onto the manifold of rank $r$ matrices via SVD after the last gradient ascent step. In all our tests, this approach produced a robust accuracy performance that is essentially the same as \ALGNAME{}, while requiring significantly more execution time (due to the additional SVD projection step), as shown in Table~\ref{table:time}. For this reason, we do not include this method in the performance comparison of Tables~\ref{table:init} and \ref{table:main}.

\paragraph{Initialization}
\def\first#1{\fcolorbox{black}{black!15}{\textbf{#1}}}
\def\second#1{\fcolorbox{black!50}{black!5}{#1}}
\begin{table*}[tb]
\caption{Comparison of robust accuracy $\rho$ across different datasets, models and algorithms. Each block corresponds to a different initialization strategy. All attacks are computed with $\|\delta X\|_2 = \tau$. Best results per model are highlighted.}
\label{table:init}
\centering
\setlength{\tabcolsep}{1.5mm} 
\resizebox{\textwidth}{!}{
\begin{NiceTabular}{cc ccccc  c ccccc c}
\toprule
 & & \multicolumn{5}{c}{\cifar{}}&  \multicolumn{5}{c}{\imagenet{}} & &\multirow{2}{*}{$\begin{array}{c}\text{Rank} \\ r=\cdot\% R\end{array}$}\\
 \cmidrule{3-7} \cmidrule{9-13} 
&  & Standard & Wa23 & Re21 & Au20 & Ri20 & & Standard & Wo20 & Li23 & En19 & Sa20 &  \\
\midrule
\multirow{5}{*}{\rotatebox[origin=c]{90}{Random}} & \multirow{5}{*}{\first{\ALGNAME}}   & 0.08  & 0.886 & 0.836 & 0.847 & 0.752 & & 0.017 & 0.355 & 0.696 & 0.472 & 0.455 & 10\%\\
& & 0.042 & 0.871 & 0.821 & 0.831 & 0.729   & &  0.01  & 0.349 & 0.692 & 0.464 & 0.445 &20\%\\
& & 0.025 & 0.864 & 0.816 & 0.824 & 0.719  & &  0.007 & 0.344 & 0.689 & 0.461 & 0.442 &30\%\\
& & 0.02  & 0.862 & 0.813 & 0.822 & 0.716   & & 0.006 & 0.342 & 0.689 & 0.462 & 0.441 &40\%\\
& & 0.017 & 0.861 & 0.813 & 0.821 & 0.717   & & 0.006 & 0.342 & 0.69  & 0.46  & 0.438 &50\%\\
\midrule
\multirow{5}{*}{\rotatebox[origin=c]{90}{Transfer}} & \multirow{5}{*}{\first{\ALGNAME}}   & 0.07  & 0.887 & 0.836 & 0.846 & 0.75 & & 0.014 & 0.356 & 0.695 & 0.469 & 0.449 & 10\%\\
& & 0.03  & 0.869 & 0.82  & 0.829 & 0.726 & &  0.006 & 0.342 & 0.69  & 0.459 & 0.438 &20\%\\
& & 0.015 & 0.861 & 0.814 & 0.821 & 0.716 & & 0.004 & 0.338 & 0.687 & 0.454 & 0.433 &30\%\\
& & 0.013 & 0.859 & 0.812 & 0.818 & 0.715 & & \first{0.003} & 0.336 & 0.687 & 0.454 & 0.432 &40\%\\
& & \first{0.009} & 0.858 & 0.81  & 0.817 & 0.714 & & 0.003 & 0.335 & 0.686 & 0.453 & 0.43  &50\%\\
\midrule
\multirow{5}{*}{\rotatebox[origin=c]{90}{Warm-up}} & \multirow{5}{*}{\first{\ALGNAME}}   & 0.07  & 0.883 & 0.833 & 0.844 & 0.747 & & 0.014 & 0.347 & 0.692 & 0.46  & 0.44  & 10\%\\
& & 0.031 & 0.867 & 0.819 & 0.828 & 0.724 & &  0.007 & 0.336 & 0.688 & 0.452 & 0.432 &20\%\\
& & 0.015 & 0.862 & 0.812 & 0.82  & 0.713 & & 0.004 & 0.33  & 0.688 & 0.447 & 0.427 &30\%\\
& & 0.013 & 0.858 & 0.81  & 0.817 & 0.712 & & \first{0.003} & 0.329 & \second{0.686} & \second{0.444} & 0.427 &40\%\\
& & 0.01  & \second{0.857} & \second{0.809} & \second{0.816} & \second{0.711} & & 0.003 & \second{0.326} & 0.686 & 0.446 & \second{0.426}  &50\%\\
\midrule
\multicolumn{2}{c}{Classic PGD}  & 0.018 & \first{0.85}  & \first{0.801} & \first{0.813} & \first{0.698} & & 0.003 & \first{0.304} & \first{0.675} & \first{0.425} & \first{0.399} & 100\%\\

 \bottomrule
\end{NiceTabular}
}
\end{table*}
\def\first#1{\fcolorbox{black}{black!15}{\textbf{#1}}}
\def\second#1{\fcolorbox{black!50}{black!5}{#1}}
\begin{table*}[tb]
\caption{Comparison of robust accuracy $\rho$ across different datasets, models and algorithms. All attacks are computed with $\|\delta X\|_2=\tau$ for classic PGD and $\|\delta U\otimes_C \delta V\|_*=\|\delta X\|_*$ for \ALGNAME.}
\label{table:main}
\centering
\setlength{\tabcolsep}{1.5mm} 
\resizebox{\textwidth}{!}{
\begin{NiceTabular}{cc ccccc  c ccccc c}
\toprule
 & & \multicolumn{5}{c}{\cifar{}}&  \multicolumn{5}{c}{\imagenet{}} & &\multirow{2}{*}{$\begin{array}{c}\text{Rank} \\ r=\cdot\% R\end{array}$}\\
 \cmidrule{3-7} \cmidrule{9-13} 
&  & Standard & Wa23 & Re21 & Au20 & Ri20 & & Standard & Wo20 & Li23 & En19 & Sa20 &  \\
\midrule

\multirow{5}{*}{\rotatebox[origin=c]{90}{$\|\|_*^{\text{\tiny LR}} = \|\|_*^{\text{\tiny PGD}} $}} & \multirow{5}{*}{\first{\ALGNAME}} & \second{0.002} & \first{0.718} & \first{0.679} & \first{0.675} & \first{0.508} & & \first{0.001} & \first{0.146} & \first{0.568} & \first{0.265} & \first{0.237} & 10\%\\
& & 0.002 & \second{0.776} & \second{0.738} & \second{0.726} & \second{0.588}   & & 0.001 & \second{0.181} & \second{0.604} & \second{0.3}   & \second{0.278} & 20\%\\
& & 0.002 & 0.808 & 0.765 & 0.753 & 0.63   & & 0.001 & 0.194 & 0.616 & 0.314 & 0.29 & 30\%\\
& & 0.002 & 0.816 & 0.774 & 0.764 & 0.643  & & 0.001 & 0.203 & 0.621 & 0.318 & 0.297 & 40\%\\
& & \first{0.001} & 0.822 & 0.779 & 0.771 & 0.651   & & 0.001 & 0.208 & 0.624 & 0.322 & 0.296 & 50\%\\
\midrule
\multicolumn{2}{c}{Classic PGD}  & 0.018 & 0.85  & 0.801  & 0.813 & 0.698 & & 0.004 & 0.304 & 0.675 & 0.425 & 0.399 & 100\%\\

 \bottomrule
\end{NiceTabular}
}
\end{table*}

We consider three types of initialization for the experiments:

\begin{itemize}
    \item \textbf{Random Initialization.} This approach follows a similar procedure to LoRA~\cite{hu2022lora}. Specifically, one of the matrices in the factorization is initialized from a Gaussian distribution, while the other is initialized with zeros. This setup ensures that the optimization procedure begins from the original image $X$.
    
    \item \textbf{Transfer Initialization.} This initialization tests the ability of \ALGNAME{} to function as a transfer learner for adversarial attacks. For a fixed dataset, we first compute the full-rank FSGM attack on the standard model and use it as the starting point for both full PGD and \ALGNAME{} on other models. For \ALGNAME{}, we additionally compute the SVD of the FSGM attack to match the desired rank for initialization.
    
    \item \textbf{Warm-up Initialization.} Similar to transfer initialization, this approach involves computing the FSGM attack, but it is tailored for each individual model and dataset. The resulting FSGM attack is then used as the starting point for both PGD and \ALGNAME{}.
\end{itemize}

It is important to note that the three initialization methods did not affect the robust accuracy performance of PGD. Instead, they had a noticeable impact on the results of \ALGNAME{}.

\paragraph{Datasets}
For the experimental part, we decided to use \cifar{} \cite{krizhevsky2009learning} and \imagenet{} \cite{deng2009imagenet}. The test sets contain 5000 examples in both cases. The former dataset has 10 classes and the latter 1000 classes. The original resolution of the images in the two datasets is really different: \cifar{} contains images whose size is $3\times 32 \times 32$, while \imagenet{} offers $3 \times 256 \times 256$ sizes images,  additionally, as a result of preprocessing the \imagenet{} size is decreased to $3 \times 224 \times 224$ for all models that are used in provided experiments. The reason for the choice of these datasets is their popularity in testing adversarial attacks.

\paragraph{Models}
For both datasets we consider a standard (non-robust) model and several robust models from Robustbench ModelZoo \cite{croce2021robustbench}. For \cifar{} the standard model uses WideResNet-28-10 architecture, for \imagenet{} -- Resnet-50 is used. To test against robust models (Wang2023 \cite{wang2023better}, Rebuffi2021 \cite{rebuffi2021fixing}, Augustin2020 \cite{augustin2020adversarial}, Rice2020 \cite{rice2020overfitting}) were chosen for \cifar{} dataset and (Liu2023 \cite{liu2023comprehensive}, Salman2020 \cite{salman2020adversarially}, Wong2020 \cite{wong2020fast}, Engstrom2019 \cite{robustness}) for \imagenet{} dataset. 

Methods are compared in terms of robust accuracy, $\rho$. This is defined through the use of the hard class function
\[
C(\delta X) = \arg \max_{j=1,\dots,d} f^{j}_\theta(X+\delta X)
\]
as follows:
\begin{equation*}
    \rho = \frac{1}{N_D}\sum_{i=1}^{N_{D}} \mathds{1}_{C(0)}(C(\delta X_i))
\end{equation*}
This metric is nothing but the percentage of dataset images for which the adversarial perturbation leaves unchanged the models' output class.

\begin{figure}[t]
\includegraphics[width=1.0\columnwidth]{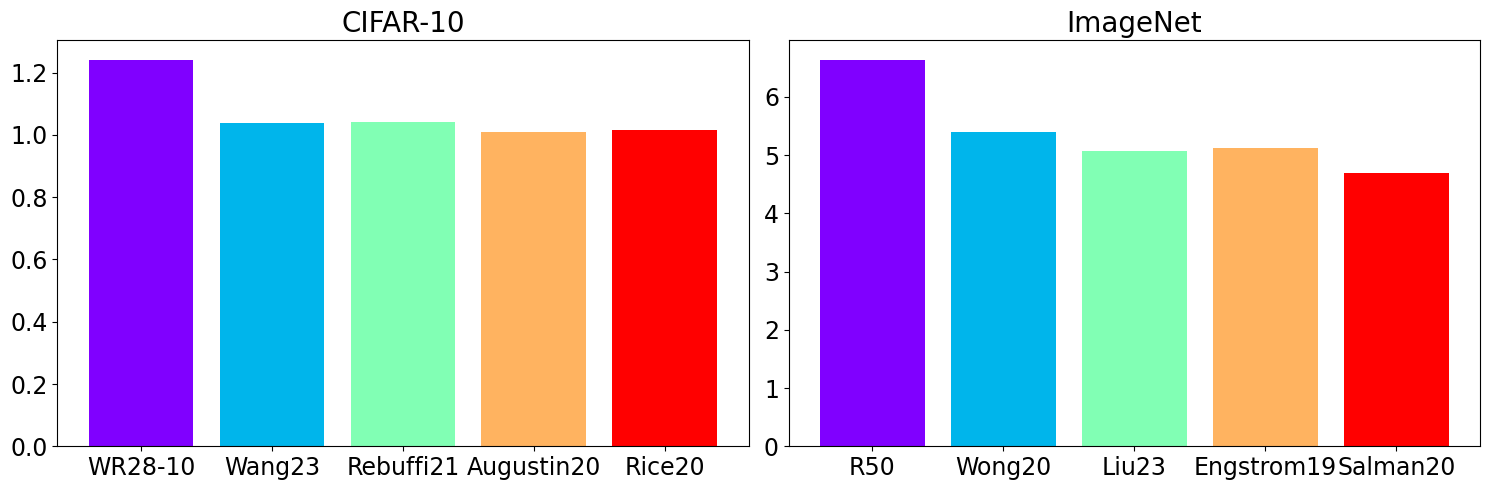}%
\caption{Nuclear norms of PGD attacks (10 steps) averaged over 5000 images from \cifar{} (left) and \imagenet{} (right) datasets.}
\label{fig:nuc_norms}
\end{figure}

\paragraph{Nuclear magnitude of adversarial attacks}

Point-wise $l^p$ norms are the typical choice for measuring adversarial perturbations. However, our observations in Figure~\ref{fig:rel_sing} suggest that concentrating the strength of the attack on specific portions of the singular spectrum could be more effective. To investigate the impact that singular values may have on the effectiveness of an attack, we measure the size of PGD attacks on both standard and robust models in terms of their nuclear norm. Figure~\ref{fig:nuc_norms} shows that, from the perspective of singular values, PGD generates stronger attacks on standard models, while the nuclear norm of the computed attacks is smaller when applied to adversarially robust models. This observation aligns with recent analyses of the frequency profiles of PGD attacks, as measured under discrete Fourier and cosine transforms \cite{yin2019fourier, maiya2021frequency}.

\begin{figure*}[t]
\includegraphics[width=.326\linewidth]{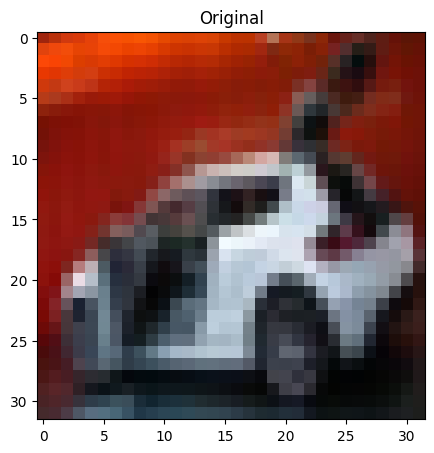}
\includegraphics[width=.326\linewidth]{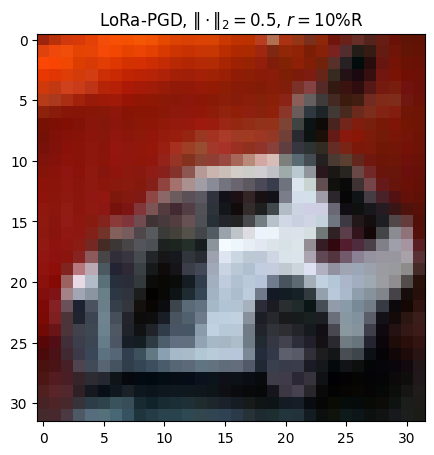}
\includegraphics[width=.326\linewidth]{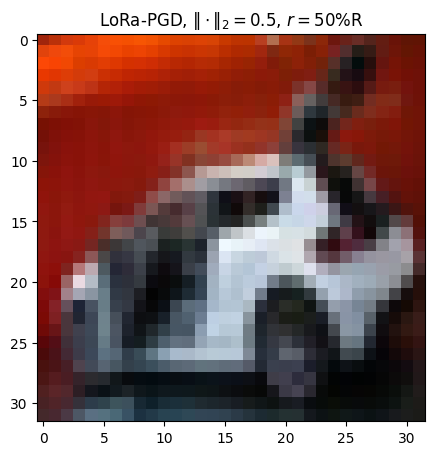} \\

\includegraphics[width=.326\linewidth]{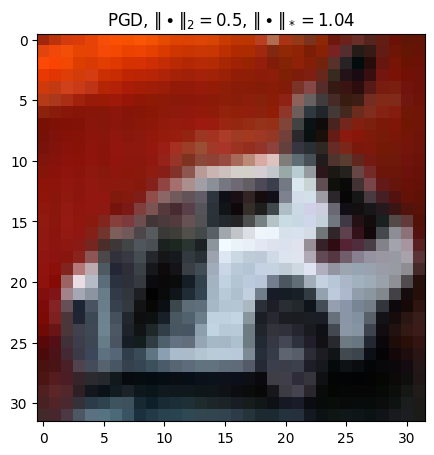}
\includegraphics[width=.326\linewidth]{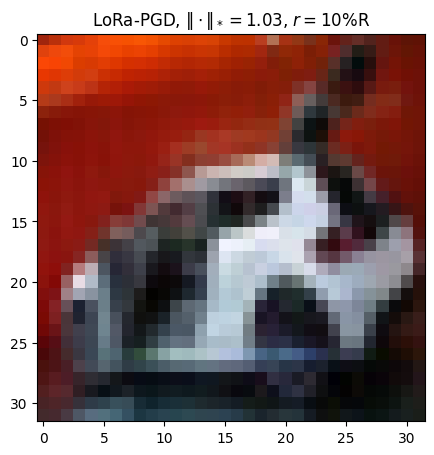}
\includegraphics[width=.326\linewidth]{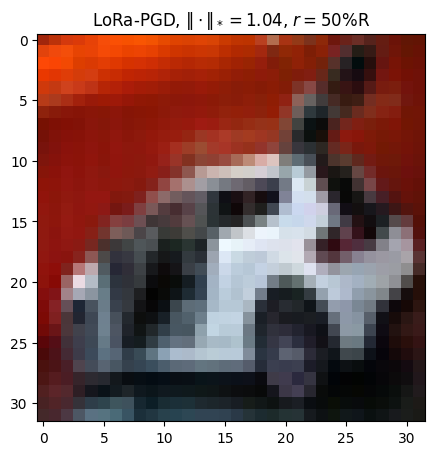}

\caption{Perceivability examples, \cifar{} dataset, Wang23 model}
\label{fig:perception_cifar}
\end{figure*}

\paragraph{Performance results}

The performance of PGD and \ALGNAME{} attacks is summarized in Tables~\ref{table:init} and \ref{table:main}. In these experiments, each attack is computed once by using $10$ gradient steps. For \ALGNAME{}, we tested relative ranks ranging from $10\%$ to $50\%$ of the maximum rank.

We set the maximum $l^2$ Frobenius norm for the PGD attack to $\tau=0.5$ for \cifar{} (a typical choice, as used in \cite{croce2021robustbench}) and $\tau=0.25$ for \imagenet{} (following the usual scaling between \cifar{} and \imagenet{} attack sizes). Table~\ref{table:init} presents results for the three initialization strategies, under the same $\tau$ values and the same $l^2$ norm for the attacks of both full rank PGD and \ALGNAME{}. In Table~\ref{table:main} we allocate the same nuclear norm budget for both PGD and \ALGNAME{}. \ALGNAME{} shows performance comparable to full-rank PGD given an equal $l^2$ Frobenius norm budget. When considering an equal nuclear norm budget, \ALGNAME{} outperforms the full-rank PGD baseline across all rank ranges.

This superiority arises because, in the full-rank case, the nuclear norm must be spread across many singular modes, whereas the low-rank perturbations computed with \ALGNAME{} allow targeting specific, more sensitive portions of the singular spectrum with relatively larger attacks. Notably, lower-rank attacks tend to be more effective on both \cifar{} and \imagenet{}.

\paragraph{Perceivability  examples}
To further explore the differences between full-rank PGD and \ALGNAME{} attacks, as well as their perceivability, we present a visual comparison in Figure~\ref{fig:perception_cifar} using images from the \cifar{} dataset. Specifically, we compare the original unperturbed image with versions perturbed by PGD where the $l^2$ norm of the attack is fixed as \( \|\delta X\|_2 = 0.5 \); \ALGNAME{} with the same $l^2$ norm \( \|\delta U\otimes_C \delta V\|_2 = 0.5 \), and for both rank=10\%R and rank=50\%R; \ALGNAME{} with the nuclear norm adjusted to match that of the full PGD attack, i.e\ \( \|\delta U\otimes_C \delta V\|_* = \|\delta X\|_* \). All methods employ 10 gradient steps on the \cifar{} dataset.

In our experimental evaluation, visual inspection reveals that, consistently across all ranks, both attacks appear similarly imperceptible to the human eye. This observation is crucial: given a fixed norm budget, concentrating the perturbation norm along a small number of singular modes does not increase the perceivability of the attack to human observers.

\paragraph{Timing Comparison}

\begin{table*}[tb]
\caption{Timing comparison results}
\label{table:time}
\centering
\setlength{\tabcolsep}{1mm} 
\resizebox{\textwidth}{!}{
\begin{NiceTabular}{ll c  cccc c cccc c}
\toprule
 & & & \multicolumn{4}{c}{\cifar{}}& & \multicolumn{4}{c}{\imagenet{}} & \multirow{2}{*}{}\\
 \cmidrule{4-7} \cmidrule{9-12} 
\multicolumn{2}{c}{Model and rank} & & St. $10\%R$ & St. $50\%R$ & Wa23 $10\%R$ & Wa23 $50\%R$ && St. $10\%R$ & St. $50\%R$ &  
Li23 $10\%R$ & Li23 $50\%R$ &   \\
\midrule
\multicolumn{2}{c}{\textbf{\ALGNAME}} & &  286.742 & 291.064 & 334.981 & 334.063 && 727.588  & 738.076 & 9963.73 & 9971.23 & \multirow{3}{*}{ms} \\
\multicolumn{2}{c}{Rank-projected PGD} & & 377.855 & 382.463 & 438.31 & 438.263  &&  3166.889 & 3167.103  & 12246.975 & 12249.441 & \\
\multicolumn{2}{c}{Classic PGD} & & \multicolumn{2}{c}{275.393} & \multicolumn{2}{c}{326.384} && \multicolumn{2}{c}{716.583}  & \multicolumn{2}{c}{9941.04} &  \\

 \bottomrule
\end{NiceTabular}
}
\vskip -.5em
\end{table*}

Fast execution time is a critical factor in selecting an attack for adversarial training. To evaluate this, we tested \ALGNAME{} against PGD and a naive low-rank projection of PGD onto the rank-\(r\) manifold, referred to as `rank-projected PGD,' using an NVIDIA A100 GPU. The results, presented in Table~\ref{table:time}, indicate that our approach is significantly faster than the naive low-rank PGD method. Furthermore, \ALGNAME{} exhibits comparable timing to PGD, with the average time per batch of 100 images being within a few milliseconds of the PGD timing.

\paragraph{Memory Comparison}

One of the major advantages of \ALGNAME{} is its lower memory cost. This reduction means that less memory is required during the computation of adversarial attacks, allowing these attacks to be precomputed and stored more efficiently for adversarial training. \ALGNAME{} leverages the $U,V$ decomposition format to store adversarial attacks, which results in significant memory savings. Specifically, for $D$ images of size $C \times M \times N$, the PGD attack requires storage in a tensor with $D C M N$ elements. In contrast, the \ALGNAME{} attack is represented by two tensors with a total of $D C r (M + N)$ elements, where $r \leq \min(M, N)$.

The memory cost difference is evident: if the image dimensions are square ($M = N$), PGD's memory usage scales quadratically with $N$, whereas \ALGNAME{}'s low-rank decomposition scales linearly with $N$ for a fixed rank $r$.

A comparison of actual memory consumption between PGD and \ALGNAME{} is provided in Table~\ref{tab:memory}. This test, conducted on 5000 examples from the \cifar{} test dataset, confirms that the observed memory usage aligns with the complexity estimation described above.

\begin{table}[t]
    \centering
    \setlength{\tabcolsep}{1mm} 
    \begin{tabular}{ccccccc}
    \toprule
         Method, $\%R$ & PGD & $10$ & $20$ & $30$ & $40$ & $50$\\
         \midrule
         Mem., Kb &  12002 & 2254 & 4504 & 7504 & 9754 & 12002\\
         \bottomrule
    \end{tabular}
    
    \caption{Memory comparison for \cifar{} dataset on WideResNet-28-10. The timing is for a batch of $5000$ images.}
    \label{tab:memory}
\end{table}

\section{Limitations}

As \ALGNAME{} shares some common features with the classical PGD attack, it also inherits certain limitations. In particular, gradient-based methods are generally ill-suited for black-box settings. Additionally, from a timing perspective, the computation of gradients imposes a lower bound on the potential speed improvements. Furthermore, while \ALGNAME{} is quite effective at lower relative ranks, its time and memory advantages diminish as the relative rank increases due to the additional computational cost associated with gradient calculations.

\section{Discussion}

In this work, we introduced \ALGNAME{}, a novel algorithm that leverages low-rank tensor structures to efficiently produce adversarial attacks. Motivated by the empirical observation that PGD tends to generate adversarial perturbations that are numerically low-rank, we utilized this insight to develop a method that produces memory-efficient attacks. At parity of the nuclear norm of the attack, \ALGNAME{} has demonstrated comparable performance to state-of-the-art methods like PGD across a variety of models and on two datasets.

Future research will explore benchmarking this method within the latest adversarial training techniques to further validate its effectiveness and practicality.

\bibliographystyle{abbrv}

\end{document}